\documentclass[lettersize,journal]{IEEEtran}
\usepackage{amsmath,amsfonts}
\usepackage{algorithmic}
\usepackage{algorithm}
\usepackage{array}
\usepackage{xcolor}
\usepackage[caption=false,font=normalsize,labelfont=sf,textfont=sf]{subfig}
\usepackage{textcomp}
\usepackage{stfloats}
\usepackage{url}
\usepackage{verbatim}
\usepackage{graphicx}
\usepackage{booktabs}
\usepackage{placeins}
\usepackage[nosort,nocompress]{cite}
\usepackage[utf8]{inputenc} 
\usepackage[T1]{fontenc}    
\usepackage{hyperref}       
\usepackage{url}            
\usepackage{booktabs}       
\usepackage{amsfonts}       
\usepackage{nicefrac}       
\usepackage{microtype}      
\usepackage{xcolor}         
\usepackage{pifont}
\usepackage{tabularx}
\usepackage{array}
\usepackage{enumitem}
\usepackage{graphicx}
\usepackage{float}
\usepackage{wrapfig}
\usepackage{caption, subcaption, overpic, textpos}
\usepackage{makecell}
\usepackage{comment}
\usepackage{amsmath}
\usepackage{amssymb}
\usepackage{multicol,multirow}
\usepackage{xspace}
\usepackage{lipsum}

\usepackage[capitalise]{cleveref}

\makeatletter
\DeclareRobustCommand\onedot{\futurelet\@let@token\@onedot}
\def\@onedot{\ifx\@let@token.\else.\null\fi\xspace}

\makeatother

\usepackage[svgnames]{xcolor}
\usepackage[table]{xcolor}

\definecolor{darkblue}{RGB}{0,50,195}
\definecolor{myblue}{rgb}{0.21,0.49,0.74} 
\hypersetup{
  pagebackref=true,
  breaklinks=true,
  colorlinks=true,
  linkcolor=red,
  urlcolor=magenta,
  citecolor=myblue,
}

\begin{document}

\title{
MM-Hand: A 21-DOF Multi-modal Modular Dexterous Robotic Hand with Remote Actuation
}

\author{
Zhuoheng Li\textsuperscript{1},
Qingquan Lin\textsuperscript{1},
Checheng Yu\textsuperscript{1},
Qiangyu Chen\textsuperscript{2},
Zhiqian Lan\textsuperscript{1},
Lutong Zhang\textsuperscript{3},
Hongyang Li\textsuperscript{1},
and Ping Luo\textsuperscript{1}%
\thanks{\textsuperscript{1}Zhuoheng Li, Qingquan Lin, Checheng Yu, Zhiqian Lan, Hongyang Li, and Ping Luo are with The University of Hong Kong, Hong Kong SAR, China.}
\thanks{\textsuperscript{2}Qiangyu Chen is with City University of Hong Kong, Hong Kong SAR, China.}
\thanks{\textsuperscript{3}Lutong Zhang is with XiFengGu Ltd., China.}
}




\maketitle

\begin{abstract}
High-DOF dexterous hands require compact actuation, rich sensing, and reliable thermal behavior, but conventional designs often occupy valuable in-hand space, increase end-effector mass, and suffer from heat accumulation near the hand. Remote tendon-driven actuation offers an alternative by relocating motors to the robot base or an external motor hub, thereby freeing the fingers and palm for additional degrees of freedom, sensing modules, and maintainable mechanical structures. This paper presents MM-Hand, a 21-DOF Multimodal Modular dexterous hand based on remote tendon-driven actuation. The hand integrates spring-return tendon-driven fingers, modular 3D-printed finger and palm structures, quick tendon connectors for maintenance, and a multimodal sensing system including joint angle sensors, tactile sensors, motor-side feedback, and in-palm stereo vision. We further analyze tendon-sheath length variation and friction loss to guide the design of the routing, motor hub, and closed-loop joint control. Experiments validate the transmission, output force, sensing, and control capability of the system. The fingertip force reaches 25N under a 1m remote sheath transmission, demonstrating practical load capacity despite long-distance tendon routing. Closed-loop joint-level experiments further evaluate command tracking with a static arm and during arm motion. These results show that MM-Hand provides a lightweight, sensor-rich, and maintainable hardware platform for dexterous manipulation research. To support the community, all hardware designs and software frameworks are made fully open-source at \texttt{\url{https://mmlab.hk/research/MM-Hand}}.
\end{abstract}

\begin{IEEEkeywords}
Dexterous Hand, Remote Tendon-Driven Actuation, Modular Robotics, Open-Source Hardware, Dexterous Manipulation
\end{IEEEkeywords}

\section{Introduction}

Dexterous hands capable of human-like in-hand manipulation remain a central challenge in robotics. The human hand combines high degree-of-freedom complexity with rich tactile sensing, enabling versatile grasping, contact transition, and fine object reorientation~\cite{jones2006human}. Reproducing even part of this capability is important for humanoid robotics, teleoperation, prosthetics, and learning-based dexterous manipulation. At the same time, dexterous hand design is constrained by several tightly coupled factors, including actuation density, end-effector inertia, sensing integration, thermal management, manufacturability, and long-term maintainability~\cite{openai2020learning,billard2019trends}.

Existing dexterous hands can be broadly categorized by transmission principle into rigid and compliant transmission. Rigid transmission, including direct-driven and linkage-driven designs, offers short transmission paths and high control accuracy~\cite{lee2016kitech,shaw2023leaphand, sharpa2025wave,psih12025official}, but embedding motors and reducers inside the hand increases mass and inertia of the hand, occupies internal volume, and worsens heat dissipation and impact robustness. Compliant transmission uses non-rigid transmission elements to convey force, including tendon-driven, artificial-muscle, and fluidic actuation. Among these, tendon-driven actuation is particularly practical for high-DoF dexterous hands because it preserves anthropomorphic morphology and spatially decouples actuators from joints, while remaining cost-effective~\cite{nazma2012tendon}. Artificial-muscle, pneumatic, and hydraulic systems offer high compliance and high force density~\cite{shintake2018soft,zhou2024dexco,phoenix2025technology}, but require demanding sealing and fabrication processes under high-load conditions.

\begin{figure}[t!]
    \centering
    \includegraphics[width=1\linewidth]{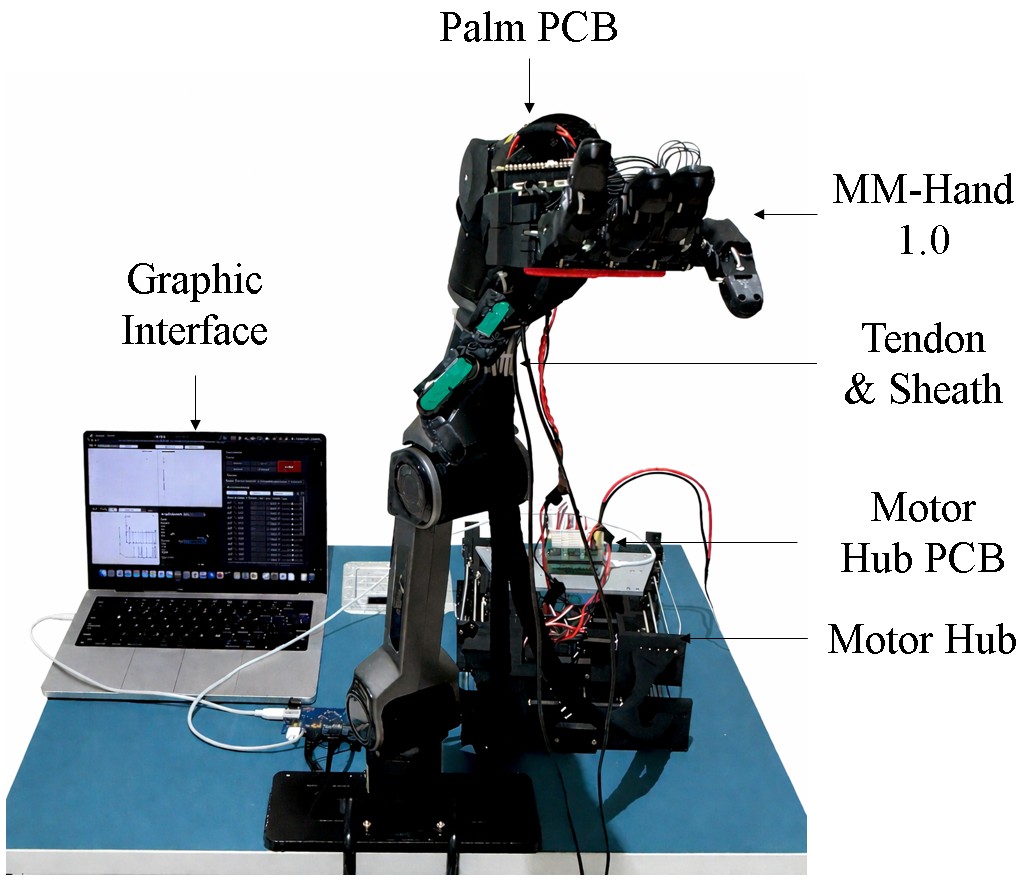}
    \caption{Overview of MM-Hand System.}
    \label{fig:teaser}
\end{figure}

Within tendon-driven systems, proximal actuation places motors in the palm or forearm, reducing finger inertia compared with fully integrated actuation and enabling effective research platforms~\cite{toshimitsu2023getting,dexhand2025paper,christoph2025orca,zorin2025ruka}. However, proximal actuation still imposes strict constraints on the payload capacity of robotic arms, heat dissipation within the hand, and actuator size. Remote tendon-driven actuation pushes this decoupling further by relocating the actuators to a standalone motor hub and transmitting power through long tendon-sheath paths to the hand~\cite{pantheon2025official,min2025antagonistic}. This architecture is attractive for lightweight humanoid and mobile platforms because it can substantially reduce hand mass and free valuable in-hand volume for sensing and structural design. However, the long tendon-sheath transmission introduces increased friction and arm-motion-dependent variations in tendon path length, both of which affect the control of finger joint angles. The tendon sheath also has a non-negligible weight.

In this work, we present \textbf{MM-Hand}, a modular and multimodal-sensing dexterous robotic hand designed for remote tendon-driven actuation. The main contributions of this paper are as follows:
\begin{itemize}
    \item We establish an analytical framework for remote tendon-driven transmission, capturing arm-motion-induced coupling disturbance and long-distance friction loss.
    \item We develop a 21-DoF, 3D-printed, modular hand structure that reserves internal space for tendon routing, sensing, and maintenance.
    \item We integrate a multimodal sensing and feedback system with joint encoders, tactile sensing, and in-palm stereo vision, forming an open and extensible hardware platform for dexterous manipulation research. We also explore tendon tension sensing in our designs.
\end{itemize}

\begin{figure*}[t!]
    \centering

    \begin{minipage}[t]{0.74\textwidth}
        \centering
        \begin{minipage}[c][5.4cm][c]{\linewidth}
            \centering
            \includegraphics[width=\linewidth,height=5.1cm,keepaspectratio]{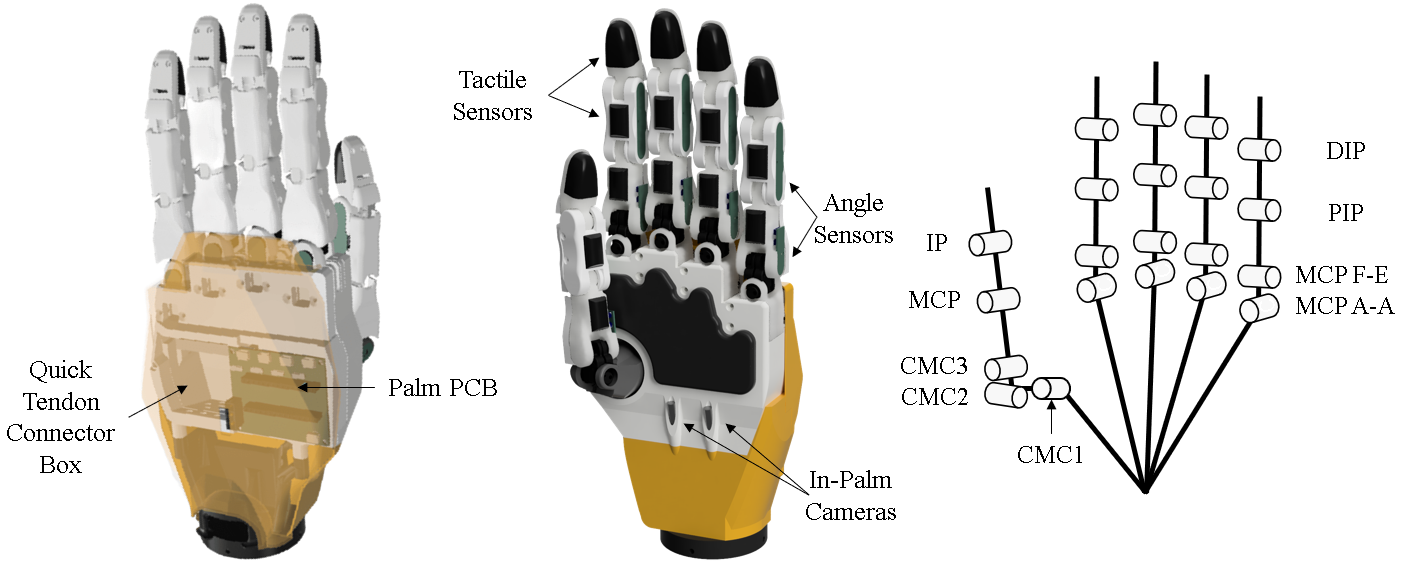}
        \end{minipage}
        \captionof{figure}{Hand structure, sensors, and degree of freedom distribution of MM-Hand. A-A denotes abduction-adduction, and F-E denotes flexion-extension.}
        \label{fig:palm}
    \end{minipage}\hfill
    \begin{minipage}[t]{0.24\textwidth}
        \centering
        \begin{minipage}[c][5.4cm][c]{\linewidth}
            \centering
            \includegraphics[width=\linewidth,height=5.1cm,keepaspectratio]{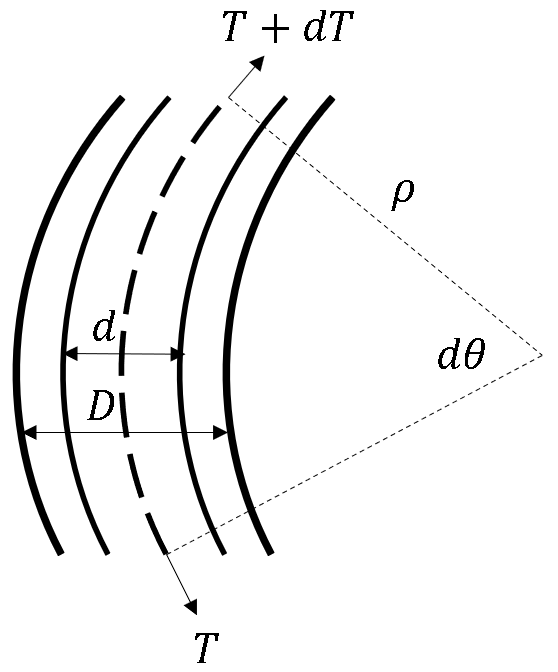}
        \end{minipage}
        \captionof{figure}{Path length and friction analysis.}
        \label{fig:Sheath-Tendon}
    \end{minipage}

\end{figure*}

\section{Theoretical Analysis of Remote Activation}

Remote tendon-driven actuation can be realized through two tendon-routing architectures. In the first, tendons are routed directly through the interior of the robotic arm. This architecture offers relatively low transmission friction and has been explored in tendon-driven arms, such as LIMS2-AMBIDEX ~\cite{song2018lims2} and recent humanoid platforms including Astribot S1 and 1X NEO~\cite{astribot2025s1,onex2025neo}. However, the tendon path length from the remote actuation unit to the hand becomes highly sensitive to robotic arm joint motion. Mitigating this coupling typically requires specially designed hollow-joint arms, limiting compatibility with general-purpose platforms.

In another architecture shown in Fig.~\ref{fig:Sheath-Tendon}, each tendon is routed through a sheath to form a Bowden-cable transmission mounted along the outer surface of the robotic arm. This approach has been adopted in remote tendon-driven hands such as the Pantheon Hand ~\cite{pantheon2025official} and ABCDL Hand ~\cite{min2025antagonistic}. It greatly reduces arm-motion-induced tendon length variation and improves platform compatibility, but introduces higher tendon-sheath friction. This section analyzes the tendon length variation and transmission friction of this architecture.

Consider a curved sheath shown in Fig.~\ref{fig:Sheath-Tendon} of length $L$, parameterized by arc length $s \in [0, L]$, with curvature $\kappa(s)$ and corresponding radius $\rho = 1/\kappa$. For an infinitesimal segment, the geometric relation satisfies $ds = \rho\, d\theta$.

Assuming the tendon remains taut and in continuous contact with the inner wall on the bending side, its centerline is offset from the sheath centerline by $e = (D - d)/2$. The corresponding tendon segment length can thus be written as $dl_t = (\rho - e)\, d\theta = (1 - e\kappa)\, ds$.

Integrating along the sheath gives the total tendon length
\begin{equation}
l_t = \int_0^L (1 - e\kappa(s))\, ds = L - e \int_0^L \kappa(s)\, ds,
\end{equation}
where $\phi = \int_0^L \kappa(s)\, ds$ denotes the accumulated bending angle.

This yields
\begin{equation}
l_t = L - \frac{D - d}{2}\,\phi.
\end{equation}

The induced joint variation is therefore
\begin{equation}
\Delta q = \frac{\Delta l_t}{r} = -\frac{D - d}{2r}\,\Delta \phi,
\label{eq:path_length}
\end{equation}
where $r$ is the effective transmission radius.

This result shows that tendon length variation depends linearly on the total bending angle $\phi$. In practice, since $\phi$ varies nonlinearly with the robotic arm configuration, this introduces configuration-dependent transmission error.

\subsection{Friction Analysis of the Tendon Inside the Sheath}
\label{sec:tendon_friction}

Consider an infinitesimal tendon segment corresponding to a turning angle $d\theta$. The change in tension direction induces a normal contact force that can be approximated as $dN \approx T\, d\theta$.

Assuming Coulomb sliding friction with coefficient $\mu$ between the tendon and the sheath, the corresponding friction force is $dF_f \approx \mu T\, d\theta$, which leads to a local tension drop $dT \approx \mu T\, d\theta$. This yields the differential relation
\begin{equation}
\frac{dT}{T} = \mu\, d\theta.
\end{equation}

Integrating over the total bending angle $\phi$ gives
\begin{equation}
T(L) = T(0)\exp(\mu \phi).
\end{equation}

The total friction-induced tension loss is therefore
\begin{equation}
F_{f,\mathrm{tot}} = |T(L) - T(0)| = T(0)\left(e^{\mu \phi} - 1\right).
\label{eq:friction}
\end{equation}

This exponential relation is consistent with the capstan-type friction model \cite{attaway1999mechanics}. It indicates that both the friction coefficient $\mu$ and the accumulated bending angle $\phi$ strongly affect transmission loss, highlighting the importance of minimizing sheath curvature and selecting low-friction materials.

\section{System Design}

\subsection{Finger Design}

\begin{figure*}[t!]
    \centering

    \begin{minipage}[t]{0.74\textwidth}
        \centering
        \begin{minipage}[c][5.2cm][c]{\linewidth}
            \centering
            \includegraphics[width=\linewidth,height=5cm,keepaspectratio]{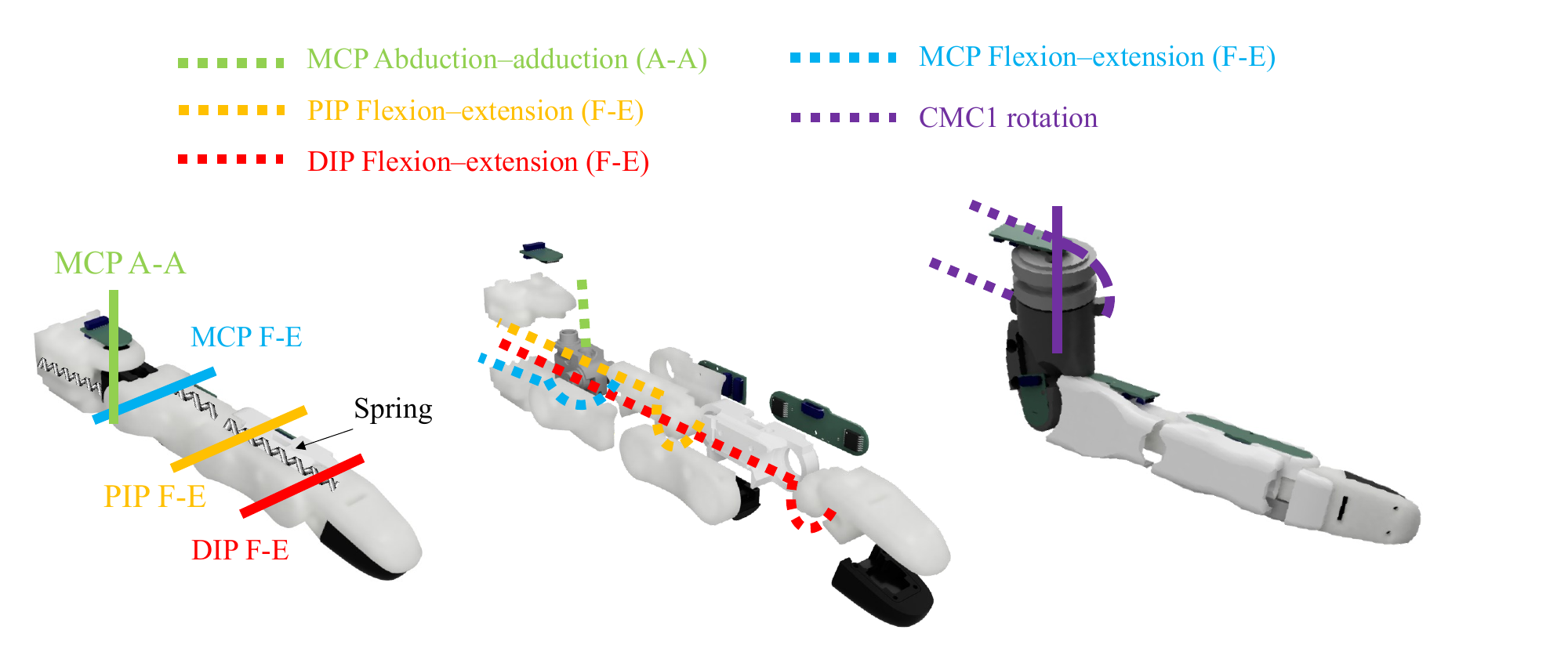}
        \end{minipage}
        \captionof{figure}{Finger structure. (a) Long finger structure with split phalanges and tendon--spring actuation. (b) Explosion view of a long finger. Dashed lines denote tendon routing paths, and solid lines denote joint-axis directions. (c) Thumb structure.}
        \label{fig:finger-structure}
    \end{minipage}\hfill
    \begin{minipage}[t]{0.24\textwidth}
        \centering
        \begin{minipage}[c][5.2cm][c]{\linewidth}
            \centering
            \includegraphics[width=\linewidth,height=5cm,keepaspectratio]{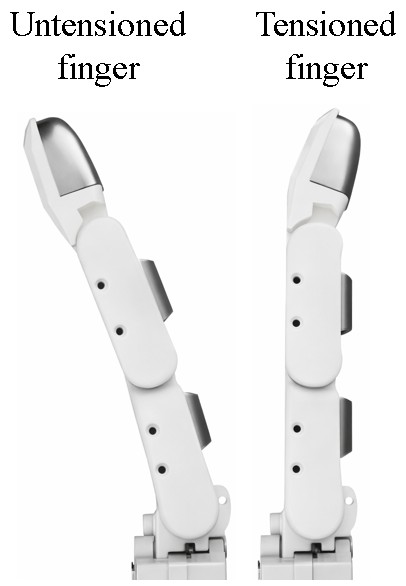}
        \end{minipage}
        \captionof{figure}{Software-based tendon pretensioning for spring-return joints.}
        \label{fig:software-tension}
    \end{minipage}

\end{figure*}

In total, MM-Hand has 21 Degree-of-Freedom (Dof) and 5 fingers. The motion range of each joint is shown in  Table.~\ref{tab:joint_range}. 

\begin{table}[t]
\centering
\caption{\textbf{Joint range of motion of the MM-Hand}.}
\label{tab:joint_range}
\small
\begin{tabular}{lccccc}
\toprule
Joint & Thumb & Index & Middle & Ring & Little \\
\midrule
CMC1 rotation & $90^\circ$ & / & / & / & / \\
CMC2 A-A      & $90^\circ$ & / & / & / & / \\
CMC3 F-E      & $90^\circ$ & / & / & / & / \\
MCP A-A       & / & $80^\circ$ & $100^\circ$ & $95^\circ$ & $85^\circ$ \\
MCP F-E       & $85^\circ$ & $90^\circ$ & $90^\circ$ & $90^\circ$ & $90^\circ$ \\
PIP F-E       & / & $85^\circ$ & $85^\circ$ & $85^\circ$ & $85^\circ$ \\
DIP F-E       & / & $90^\circ$ & $90^\circ$ & $90^\circ$ & $90^\circ$ \\
IP F-E        & $90^\circ$ & / & / & / & / \\
\bottomrule
\end{tabular}
\end{table}

The four long fingers (index, middle, ring, and little) share an identical modular architecture to simplify fabrication, assembly, and replacement. Each finger provides four DoFs, including an abduction / abduction joint of MCP and three flexion joints at MCP, PIP, and DIP. All joints adopt a single-tendon, spring-return actuation scheme, in which tendon pulling generates flexion and passive elastic elements provide extension.

As shown in Fig.~\ref{fig:finger-structure}, each finger uses a vertically split multi-piece phalanx design. The distal phalanx is divided into two longitudinal parts, while the intermediate and proximal phalanges are each divided into three parts and fastened by bolts. This design improves manufacturability for 3D printing by allowing better print orientation for smoother internal tendon channels, while also facilitating tendon installation and internal routing of sensor wires.

The thumb is designed as the main source of opposition and provides five DoFs. Its distal, middle, and proximal phalanges reuse the same split modular construction as the long fingers, while the thumb base is redesigned to accommodate an additional basal rotation DoF. Unlike other joints, multiple upstream tendons cannot be routed through the thumb base center (CMC1 joint), leading to larger parasitic torques. Thus, antagonistic dual-tendon actuation instead of a spring-return mechanism is used in this joint, as shown in Fig.~\ref{fig:finger-structure}.

\subsection{Hand Design}

The palm adopts a two-part enclosure with a top cover and a bottom cover, as shown in Fig.~\ref{fig:palm}. An in-palm quick tendon connector box is integrated into the bottom cover to centrally manage all tendon lines from the fingers and the remote motor hub. By separating the hand-side and arm-side tendon segments through detachable connectors, this design enables rapid finger or sheath replacement without re-routing the entire transmission, thereby improving modularity and maintainability.

\subsection{Motor Hub and Tendon Tension}

For spring-return joints, Eq. \eqref{eq:friction} shows that tendon–sheath friction can be exponentially amplified toward the motor-hub side during release. To ensure reliable spring-driven reset, the motor hub minimizes additional hub-side resistance: each tendon leaves the spool tangentially and enters the sheath directly, without extra bends or redirection inside the hub.

We also explored two hardware tensioning and tension-sensing mechanisms in earlier prototypes, including a pulley-based tensioner and a concentric dual-spool design. However, both increased assembly complexity and were not suitable for an open-source, easily reproducible platform. MM-Hand therefore adopts software-based pretensioning (Fig. ~\ref{fig:software-tension}): each spring-return joint is initialized with a predefined pretension angle, and the motor reels the tendon back whenever the measured joint angle indicates slack. 
This removes dedicated tensioning hardware while maintaining basic tendon engagement during the operation.

\subsection{Sensors and Electronics}


\begin{figure}[t!]
    \centering
    \includegraphics[width=.45\textwidth]{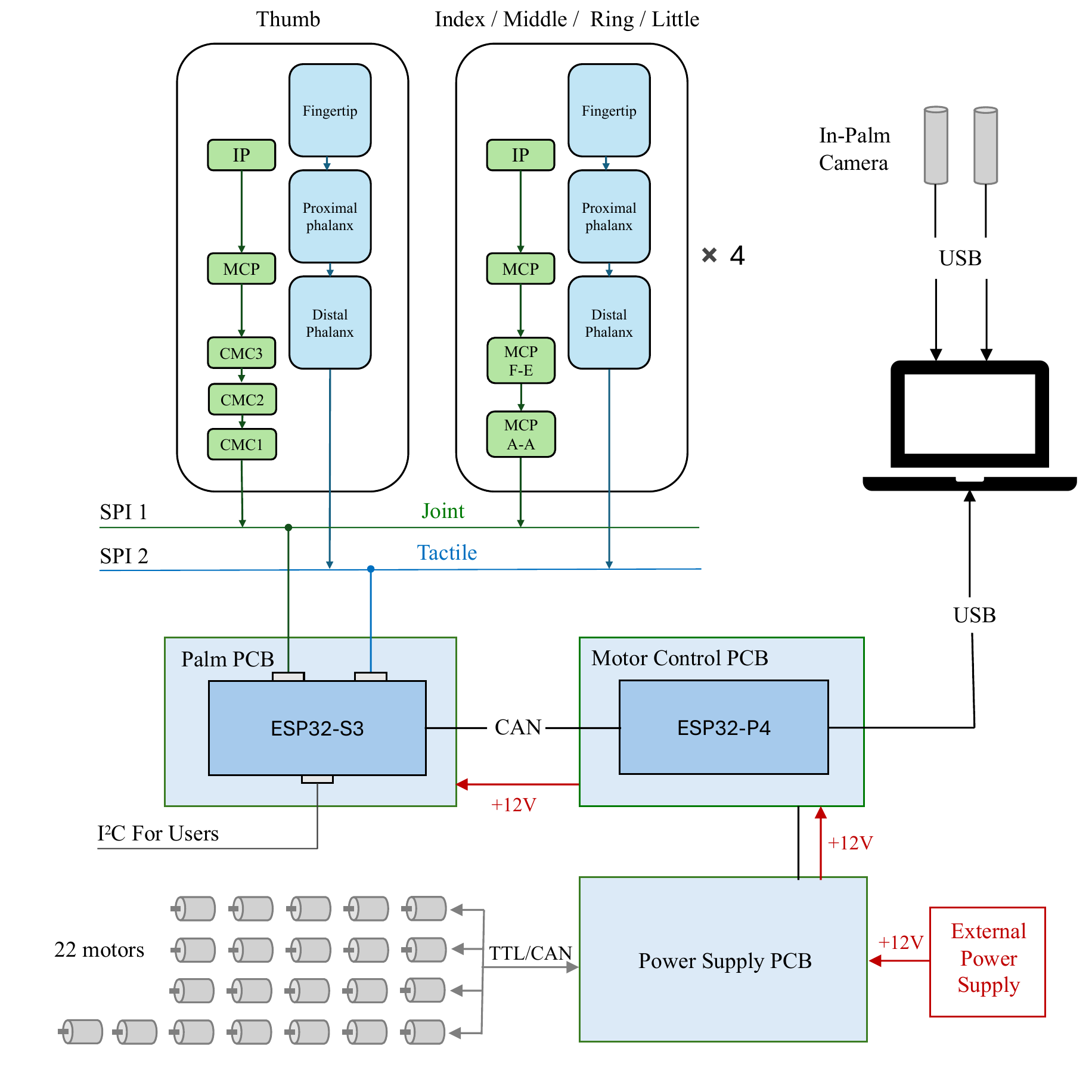}
    \caption{System-level electronics architecture of MM-Hand, including multimodal sensing (joint encoders, tactile sensors, and in-palm cameras), SPI-based finger sensor buses, palm and motor control PCBs, and CAN/USB communication interfaces.}
    \label{fig:ee}
\end{figure}

MM-Hand integrates joint angle sensing, tactile sensing, in-palm stereo vision, and motor-side encoder feedback for multimodal perception. As shown in Fig. ~\ref{fig:ee}, each finger carries joint encoders and tactile sensors. Joint angles are measured by AMS AS5047P encoders, while tactile sensing uses PaXini Elite sensors, with S2015 at the fingertips and S1610 at the phalanges.

To reduce wiring complexity, the finger sensors are organized into two shared SPI buses, one for encoders and one for tactile sensors, and all buses are aggregated on the palm PCB. Although SPI is not typically used in this manner, the encoders are connected through an SPI daisy chain, while the tactile sensors share a managed SPI bus with dedicated chip-select control.

The palm PCB collects sensor data and sends them via CAN to the motor control PCB. Motor-side encoders provide additional feedback, and the system communicates with the host computer through USB. This design enables synchronized multimodal sensing with low wiring overhead and good scalability for high-DoF dexterous hands.

\subsection{Feedback Control}

MM-Hand is driven by long tendon transmissions, where most joints use serial-bus motors (e.g., Feetech ST3215-12V or other TTL/CAN-based actuators) with passive spring return, while thumb rotation is driven by an antagonistic tendon pair. Compared with short-tendon systems, this architecture introduces stronger transmission nonlinearity and uncertainty.

As described in Eq. \eqref{eq:path_length}, the total sheath bending angle varies nonlinearly with the robotic arm pose, leading to a configuration-dependent mapping between motor rotation and joint motion. Friction and tendon stretch further introduce hysteresis and disturbance, making accurate feedforward modeling impractical.

We therefore adopt a sensor-based closed-loop strategy using joint-mounted absolute encoders. Control is performed directly in joint space, with the encoder error used to compensate for transmission loss and hysteresis through PID control. To avoid tendon slack during rapid release, the motor release speed is limited so that it does not exceed the spring recovery rate. In addition, motor current is monitored for abnormal load detection. If a current spike is inconsistent with encoder motion, the system immediately stops to prevent tendon failure or structural damage.

\section{Experiments}

We evaluate MM-Hand through several groups of experiments: sheath-tendon friction measurement, single-finger fingertip force measurement, control accuracy, and in-palm stereo vision. These experiments are designed to validate the core properties of the system, including transmission loss in the remote actuation mechanism, output capability, control robustness, and sensing functionality. We use Dyneema tendon with 1mm diameter in our setting.

\subsection{Sheath-Tendon Friction Measurement}

\begin{figure}
    \centering
    \includegraphics[width=.7\linewidth]{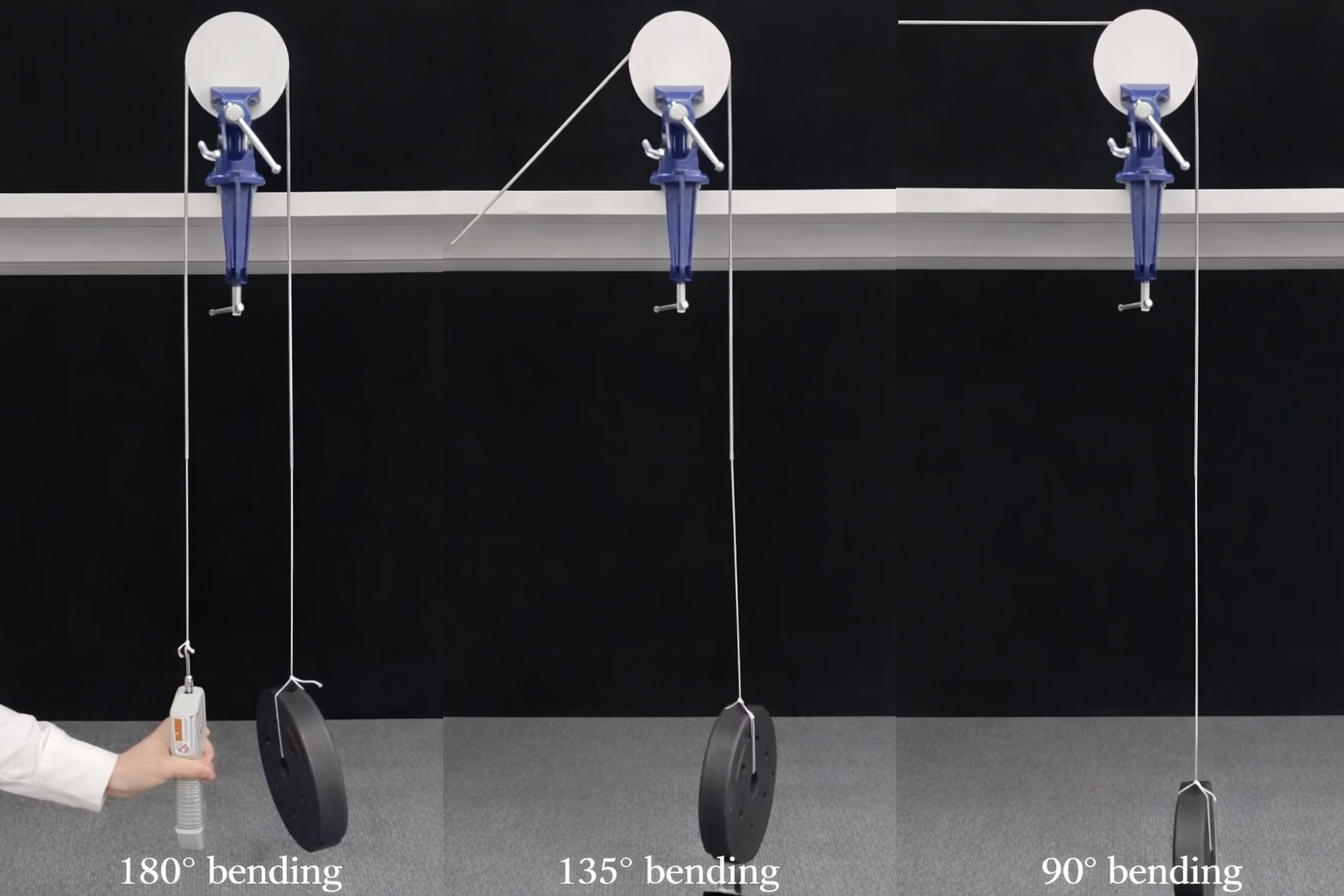}
    \caption{Setup of Sheath-Tendon Friction Measurement. }
    \label{fig:friction_test_setup}
\end{figure}

\begin{figure}
    \centering
    \includegraphics[width=.7\linewidth]{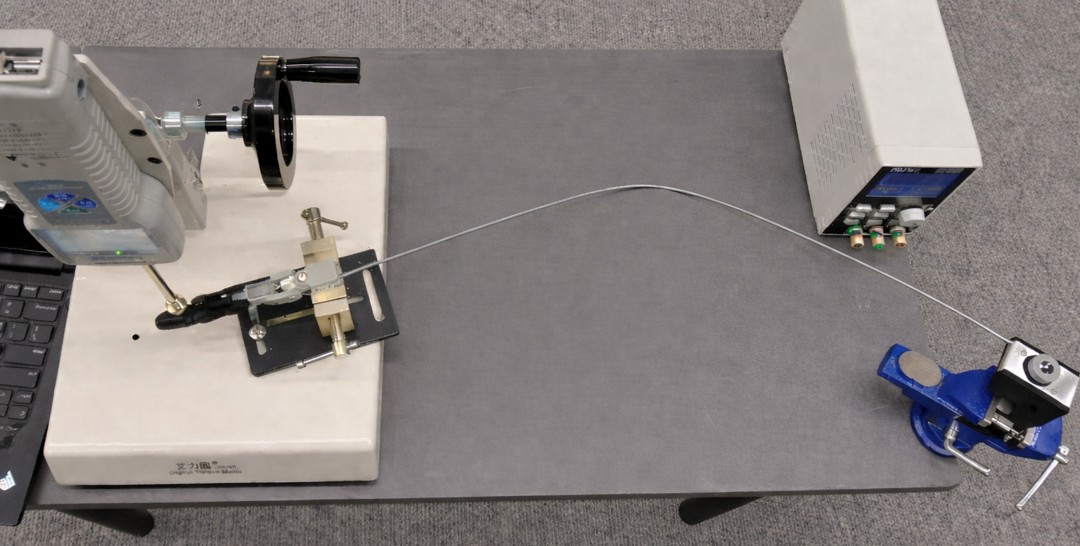}
    \caption{Setup of Fingertip Force Measurement. }
    \label{fig:fingertip_test_setup}
\end{figure}

The friction test setup is shown in Fig.~\ref{fig:friction_test_setup}. One end of the tendon was connected to a handheld force sensor, while the other end was loaded by a 2~kg weight plate. The tendon was routed around circular disks to impose wrap angles of $0^\circ$--$180^\circ$ and disk diameters of 10--100~mm. For each test condition, the tendon was pulled at an approximately constant speed and the quasi-steady portion of a single force-gauge trace is recorded. The mean friction was obtained by subtracting the 2~kg load from the mean measured tension.

\begin{figure}
    \centering
    \includegraphics[width=\linewidth]{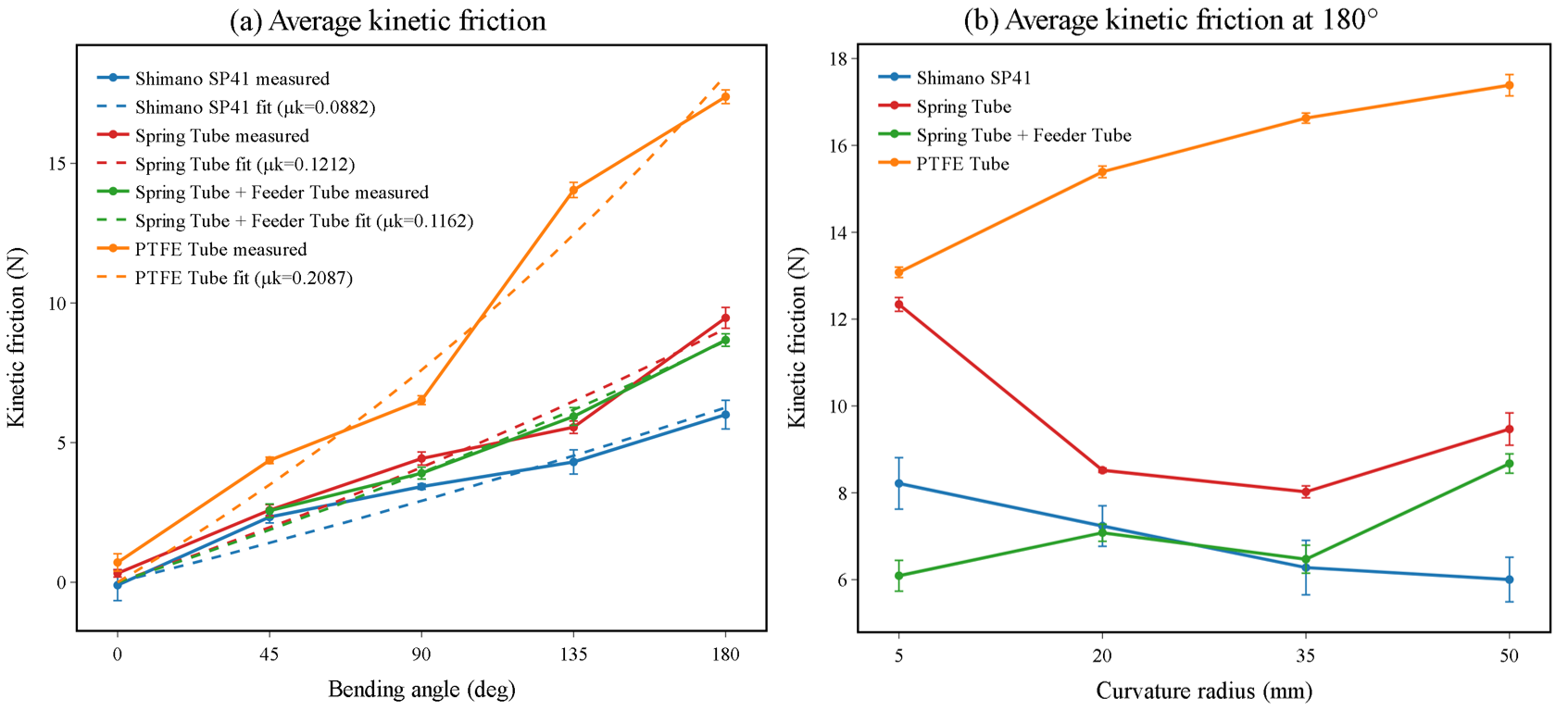}
    \caption{Friction test results for the four sheath types. (a) Average kinetic friction versus bending angle with disk diameter 100mm. Dashed lines indicate exponential fits based on Eq. \eqref{eq:friction}. Error bars denote standard variations. (b) Average kinetic friction at a wrap angle of $180^\circ$ versus curvature radius.}
    \label{fig:friction_results}
\end{figure}

Four sheath types were tested: Shimano SP41 lubricated shift housing, a standard PTFE tube, a metal spring tube, and a Bambu printer feeder tube reinforced by an outer spring sheath. Their dimensions are listed in Fig.~\ref{fig:friction_results}.

As shown in Fig.~\ref{fig:friction_results}, kinetic friction generally follows the exponential trend predicted by Eq.~\eqref{eq:friction}. Although PTFE is intrinsically low-friction, the standard PTFE tube exhibited relatively high friction, likely due to ordinary extrusion quality and inner-surface finish. The Bambu feeder tube is also PTFE-based but uses a higher-quality extrusion process, resulting in lower friction. The metal spring tube showed similarly low friction, likely because its corrugated structure reduces tendon contact to discrete points. Although Shimano SP41 had low kinetic friction but usually shows high static friction, which is not desirable for control. Thus, the metal spring tube was selected for subsequent experiments for its simplicity and balanced performance.

\subsection{Fingertip Load Performance}

Fig.~\ref{fig:fingertip_test_setup} shows the single-finger fingertip force test. The finger was fixed on a rigid fixture, with the fingertip pressed against a push--pull force gauge. The base joint was driven by a servo through a metal spring sheath. Two sheath lengths, 1~m and 0.1~m, were tested to compare remote and proximal actuation.

\begin{figure}
    \centering
    \includegraphics[width=.8\linewidth]{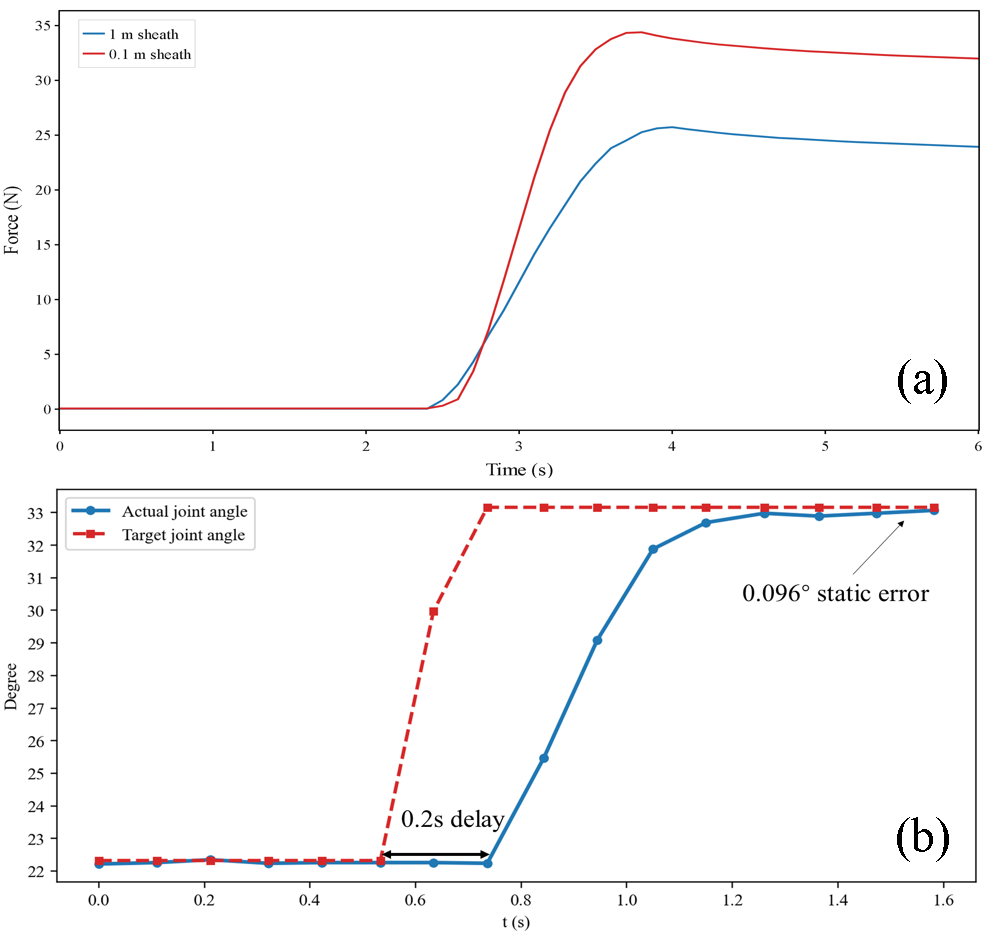}
    \caption{(a) Fingertip Force Measurement. (b) Measured joint angle static error given a step input. }
    \label{fig:fingertip_static_combine}
\end{figure}

As shown in Fig.~\ref{fig:fingertip_static_combine} (a), the 0.1~m sheath reached a higher peak force of about 33~N, compared with about 25~N for the 1~m sheath. The 8~N difference mainly reflects the additional friction loss in the longer and slightly curved sheath. In both cases, the force gradually decreased after the peak, likely due to slow sheath deformation under load while the servo command remained fixed.

\subsection{Controlling Accuracy}

Each DoF is equipped with a joint encoder for real-time feedback, and the measured joint angle is used in a PID-based closed-loop motor position controller. We evaluate the controller using two tests. In the step-response test, Joint 0 is commanded by a step input to assess static positioning accuracy. As shown in Fig.~\ref{fig:fingertip_static_combine} (b), the proposed controller achieves a steady-state error below 0.1°, while the tendon–sheath friction introduces an approximately 0.2 s delay between the onset of motor actuation and the actual onset of joint motion. 

We further evaluate dynamic tracking using a 0.5hz sinusoidal reference input on Joint 0 with over 20 s. The robotic hand is mounted on a 6-DoF Agilex Piper manipulator. In one setting, the manipulator remains stationary; in the other, it simultaneously executes a circular trajectory through the workspace to introduce transmission disturbances. The purpose of this experiment is to evaluate the effect of manipulator-induced disturbances on tendon-driven motion, and consequently on joint angle tracking performance. As shown in Fig. 12, a similar delay is observed in sinusoidal tracking. Although the tracking accuracy degrades under manipulator motion, the results show that friction-introduced delay has more impact on controlling performance. 

\begin{figure}[t]
    \centering
    \includegraphics[width=.8\linewidth]{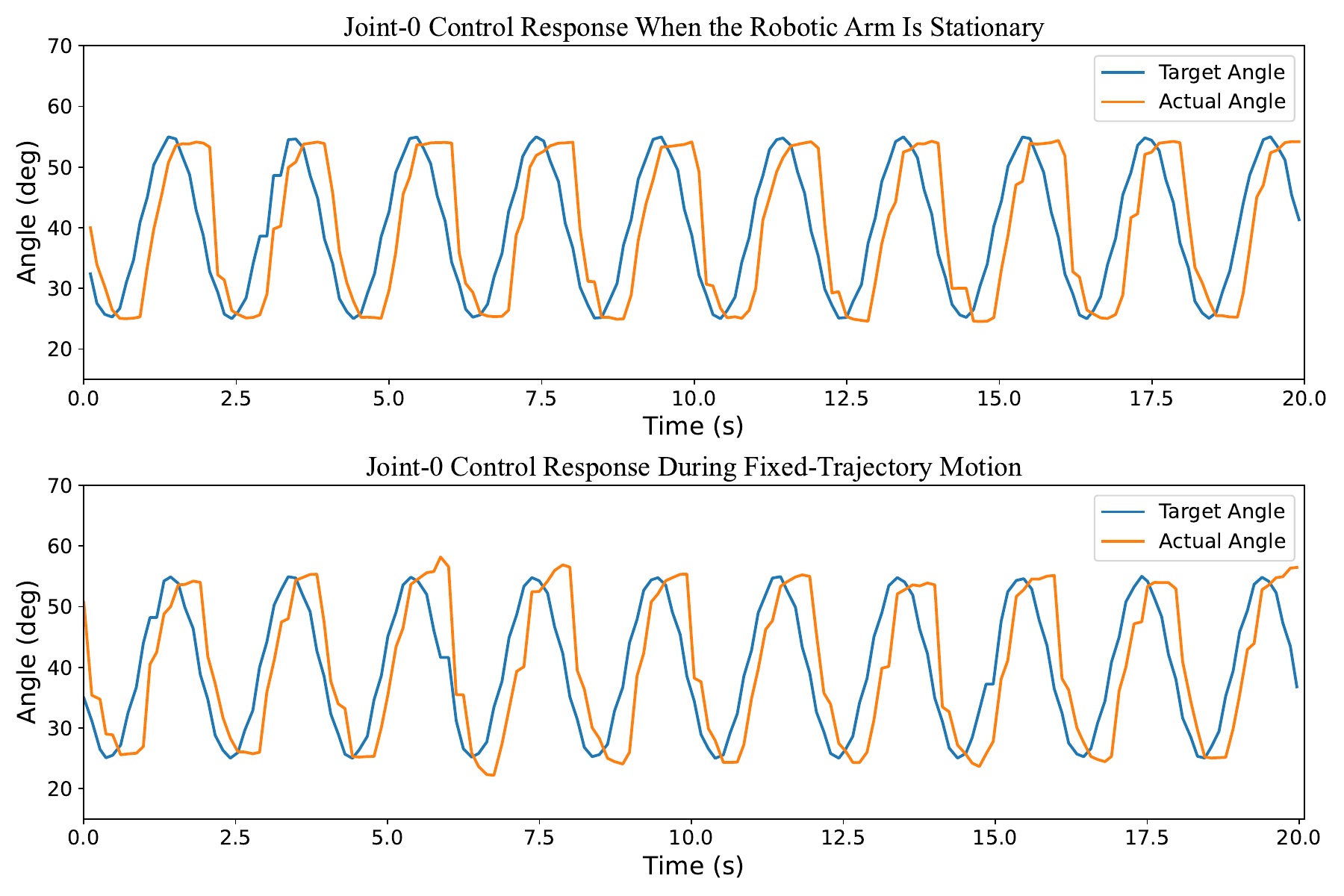}
    \caption{Tracking performance of Joint 0 under a 0.5hz reference input with amplitude ranging from $25^\circ$ to $55^\circ$.  (a) The manipulator remains stationary. (b) The manipulator executes motion throughout the workspace.}
    \label{fig:control_response}
\end{figure}

\subsection{In-Palm Stereo Depth Sensing}

\begin{figure}
    \centering
    \includegraphics[width=.7\linewidth]{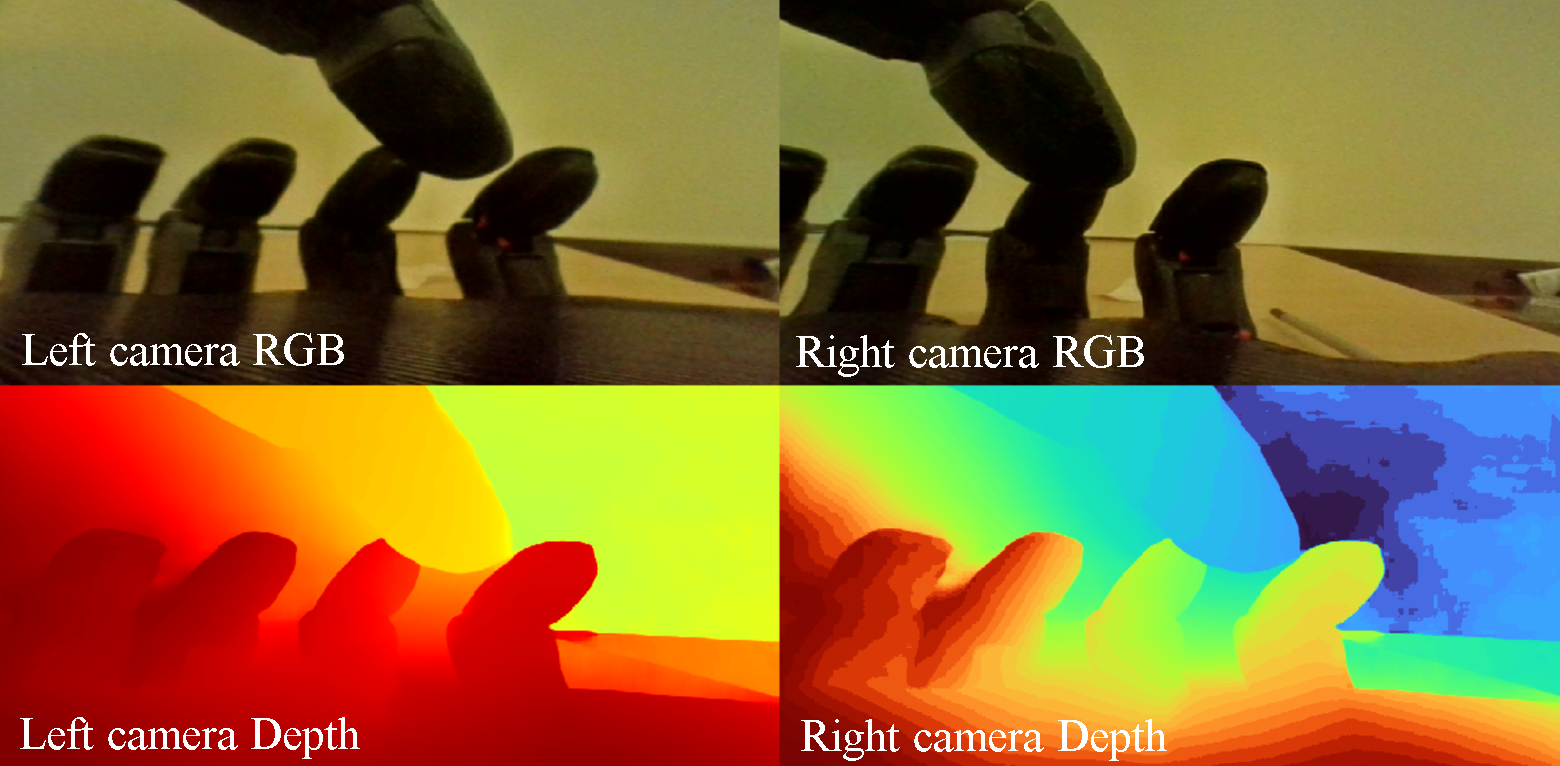}
    \caption{Stereo depth estimation from the in-palm cameras.}
    \label{fig:stereo_depth}
\end{figure}

In parallel, the in-palm stereo camera pair provides dense depth sensing during manipulation. A pretrained RAFT-Stereo network~\cite{lipson2021raft} estimates disparity from the rectified image pair, which is converted to metric depth via the calibrated baseline and focal length. The result is shown in Fig.~\ref{fig:stereo_depth}.

\section{Conclusion}

This paper presented MM-Hand, a 21-DoF remote tendon-driven dexterous hand with modular mechanics and multimodal sensing. Theoretical analysis guided the tendon-sheath routing and control design, while experiments validated friction behavior, fingertip force, sensing, and joint control. 

\textbf{Discussions \& Limitations.} Current limitations include delay because of tendon friction, tendon wear reliability, insufficient spring-return force for abduction-adduction joints, and a still-large palm size. An updated design, MM-Hand 1.1, is being tested to address these issues.


 
%

\FloatBarrier
\bibliographystyle{IEEEtran}
\bibliography{main}


 


\vfill

\end{document}